\documentclass[11pt]{article}

\usepackage[margin=1in]{geometry}  % Lots of layout options.  See http://en.wikibooks.org/wiki/LaTeX/Page_Layout
\usepackage{setspace}  % control line spacing in latex documents
\usepackage{caption}
\usepackage{gensymb}

\usepackage[fleqn]{amsmath}  % latex math
\usepackage{amssymb}
\usepackage{empheq} % http://www.ctan.org/pkg/empheq
\usepackage{bm,upgreek}  % allows you to write bold greek letters (upper & lower case)

\usepackage{changepage}

\usepackage{url}

\usepackage{algorithmic,algorithm}  

\usepackage{graphicx}  % inclusion of graphics; see: http://en.wikibooks.org/wiki/LaTeX/Importing_Graphics
\usepackage{xspace}

\usepackage{color}  % http://en.wikibooks.org/wiki/LaTeX/Colors

\usepackage{tikz}
\usetikzlibrary{fit,positioning}
\usetikzlibrary{bayesnet}
\usepackage{wrapfig}

\usepackage{footmisc}
\newcounter{secnum}
\usepackage[all]{nowidow}

\usepackage{titlesec}
\titleformat*{\section}{\Large\bfseries}
\titleformat*{\subsection}{\onehalfspacing\Large}
\titleformat*{\subsubsection}{\large\bfseries}
\titleformat*{\paragraph}{\large\bfseries}
\titleformat*{\subparagraph}{\large\bfseries}

\usepackage{fancyhdr}
\usepackage{indentfirst}
\usepackage{lipsum}  

\usepackage{adjustbox}
\let\oldhash\#%
\DeclareRobustCommand{\#}{\adjustbox{valign=B,totalheight=.35\baselineskip}{\oldhash}}%

\newfont{\namefont}{cmr10 at 12.5pt}
\usepackage[hidelinks]{hyperref}

\begin{document}
\vspace*{0cm}
\begin{center}
\noindent{\LARGE Embodied, Situated, and Grounded Intelligence: Implications for AI\\[.25cm]  (Workshop Report)}
\vspace{.5cm}

\rule{0.75\textwidth}{.4pt}
\vspace{.5cm}

\begin{minipage}{.3\textwidth}
\centering
\onehalfspacing
{\namefont Tyler Millhouse}\\ Santa Fe Institute\\ tyler.millhouse@santafe.edu\\
\end{minipage}% This must go next to `\end{minipage}`
\begin{minipage}{.3\textwidth}
\centering
\onehalfspacing
{\namefont Melanie Moses}\\ University of New Mexico\\ melaniem@cs.unm.edu\\
\end{minipage}
\begin{minipage}{.3\textwidth}
\centering
\onehalfspacing
{\namefont Melanie Mitchell}\\ Santa Fe Institute\\ mm@santafe.edu\\
\end{minipage}\vspace{.5cm}
%\noindent Tyler Millhouse, Melanie Moses, \& Melanie Mitchell

\rule{0.75\textwidth}{.4pt}
\vspace{.5cm}
\begin{quote}
\textbf{Abstract:} In April of 2022, the Santa Fe Institute hosted a workshop on embodied, situated, and grounded intelligence as part of the Institute's Foundations of Intelligence project.  The workshop brought together computer scientists, psychologists, philosophers, social scientists, and others to discuss the science of embodiment and related issues in human intelligence, and its implications for building robust, human-level AI. In this report, we summarize each of the talks and the subsequent discussions. We also draw out a number of key themes and identify important frontiers for future research.

\end{quote}

\end{center}

\newpage
\tableofcontents
\newpage
\section{Overview}

All intelligent systems, whether natural or artificial, are realized in the physical world. This entails that all intelligent systems will, in a minimal sense, have a physical body, be situated in a particular environmental context, and have concepts that are at least in part grounded in the physical world. The deeper question is whether these aspects of intelligent systems play a substantive role in intelligence itself. For example, how important is the ability of human children to explore their environment, manipulate objects, and observe the consequences of their behavior? Can we realistically expect systems without these capabilities (e.g., deep neural networks that learn passively from large data sets) to achieve robust, generalizable competence? Answering such questions requires a better understanding of intelligence itself, an understanding grounded in insights from multiple scientific disciplines and in experiments involving multiple cognitive domains. 

For this reason, embodied intelligence has become a major area of research across several fields of study, including artificial intelligence, philosophy, neuroscience, and psychology. In this workshop, we brought together experts in these fields to share their insights, learn from others, and discuss the future of embodiment research. In addition to the questions noted above, the workshop also addressed many other questions, including: What do terms such as ``embodiment'', ``situated intelligence,'' and ``grounded cognition'' mean from the perspectives of different disciplines? To what degree are our concepts and language embodied, situated, and grounded? Could machine intelligence be developed without such embodiment and grounding? How can insights from cognitive science inform important questions in AI, such as how to assess a system's ``understanding'' and generalization abilities? How can these ideas help us to make progress on socially beneficial AI, and the often discussed (but ill-defined) notion of ``AI alignment with human values''? In what follows, we summarize the insights offered by our speakers and attempt to distill the larger themes present in the talks and discussions.

\newpage
\section{Summaries of Talks and Discussions}

\subsection[``Why Intelligent Reasoning Must Be Embodied'' (Lisa Miracchi)]{``Why Intelligent Reasoning Must Be Embodied''}
\begin{adjustwidth}{1cm}{1cm}
\onehalfspacing
\textit{Lisa Miracchi is an Associate Professor of Philosophy at the University of Pennsylvania. She is also affiliated with the General Robotics, Automation, Sensing, and Perception (GRASP) Lab and MindCORE.}
\end{adjustwidth}

Lisa Miracchi laid out a philosophical argument for the importance of embodied cognition and the limitations of traditional AI approaches. Her argument is based on her commitment to ``semantic efficacy,'' the view that the meaning (or content) of mental states makes a causal difference to what agents do and how they affect their environments. For example, that to explain why an agent has taken a particular course of action, it is often helpful to appeal to what that agent believes about the world (e.g., an agent's belief that it would rain causally contributed to her picking up the umbrella). Miracchi's argument also appeals to semantic externalism---the view that the meaning or ``content'' of a mental state depends on how one is situated in one's environment. Together, these positions suggest that to understand how cognition shapes the behavior of agents, one must think about how agents are situated in their environments. 

Miracchi argues that traditional approaches to AI and cognitive science have undersold the importance of embodiment and situatedness. These approaches often combine a content-neutral description of cognitive mechanisms with ascriptions of meaning to elements of that description. The former describes the operation of the mechanism at a physical, mechanical, or (more commonly) algorithmic level. The latter expresses the significance of elements of that description in light of how the mechanism figures in an agent's cognitive economy and how the agent is embedded in its environment. To better appreciate this distinction, suppose that similar bits of neural circuitry are used to detect flies in one species of frog and ants in another, with each circuit outputting similar signals when a detection is registered. No matter how physically, mechanically, or algorithmically similar these circuits/signals are, in one species the signal means ``fly detected'' and in the other it means ``ant detected.'' This is because of the cognitive, behavioral, and environmental significance of the signal in each species. Hence, it seems like we could give a narrow, content-neutral description of how each circuit operates (e.g., identifying the algorithm it uses to processes inputs and produce outputs), then ascribe a meaning to its output based on wider considerations. 

However, Miracchi contends that this approach amounts to a rejection of semantic efficacy, since it suggests that the physical, mechanical, or algorithmic features of a cognitive system determine how it operates, and it treats the meaning of the output signal as a mere interpretive gloss,  contextualizing but not influencing the operation of the circuits. With some additional biological assumptions, we might even imagine transplanting circuits between members of each species, without making a functional difference to the subjects' thinking or behavior. On the contrary, Miracchi holds that the causal role of mental content is richer than this approach suggests. Mental content is better understood as a consequence of coordinating a diversity of internal mechanisms and external processes. It is not as if animals represent the presence of a fly or ant in their environment using a single signal or activation pattern (as we might use a single Boolean variable in a computer program). Rather, animals represent elements of their environment using multiple signals and mechanisms that are richly sensitive to the details of the present context and to feedback from the environment. Further, evolved systems are sufficiently robust that when one mechanism or signal fails, others can step in to compensate, and representations with relevantly similar contents (e.g., ``fly detected'') will differ within and between contexts (e.g., with viewing angle, illumination, feedback to other sense modalities, etc.).

On Miracchi's view, to ascribe mental contents to many mental representations only makes sense in the context of a broader capacity for coupling the states of one's internal cognitive mechanisms to the states of one's environment. To account for this capacity we must appeal to features of both our cognitive mechanisms \textit{and} the elements of our environment to which we are coupled. As such, we cannot hope to characterize all cognitive processes in the abstract (i.e.,  without considering those elements of the world to which they are related) since cognitive processes involving brain-environment coupling depend on causal interactions between internal mechanisms and the wider environment. This makes the referents of our mental representations an active part of ongoing cognition and not merely part of an interpretative gloss of an abstractly characterized cognitive mechanism. Miracchi argues that nothing about this view ascribes magical powers to meanings, but rather explains how the external factors which fix the meaning of our mental representations can play an active role in cognition. 
\begin{center}
\textit{To watch the entire talk, please visit: \href{https://youtu.be/eftmRJwkMCQ}{https://youtu.be/eftmRJwkMCQ}}  
\end{center}
\noindent\textbf{Discussion:} 

In the discussion section, participants were divided about the importance of appealing to the causal efficacy of mental contents.  Two commenters argued that it is precisely the ability of cognitive systems to act \textit{as if} meanings are causally efficacious that makes these systems so interesting. The physical stuff of brains, as such, doesn't really have any semantic contents, but it is so subtly arranged that it behaves as if it does. This approach does not deny the importance of mental content ascriptions in interpreting and predicting the behavior of cognitive systems, but it emphasizes that the design challenge of developing such systems (i.e., the challenge we face in AI) is precisely the challenge of getting meaningless mechanisms to exhibit this kind of sophistication. Miracchi responded that some aspects of reasoning require an appeal to higher-level phenomena involving agent-environment interactions and that solving the problem of designing cognitive systems requires us to imagine how we might realize systems that exhibit these higher levels of organization and behavior. For example, thermal regulation in animals involves a host of mechanisms coupled to environmental feedback, no one of which is necessary or sufficient for thermal regulation. To attribute a capacity for thermal regulation to animals, then, cannot appeal to causal powers of any particular mechanism but must instead appeal to higher-level facts about the organization of the thermal regulation systems and its coupling to the environment.

\subsection[``Embodied Cognition: Some Philosophical Proposals'' (Kenneth Aizawa)]{``Embodied Cognition: Some Philosophical Proposals''}
\begin{adjustwidth}{1cm}{1cm}
\onehalfspacing
\textit{Kenneth Aizawa is a Professor of Philosophy at Rutgers University. He specializes in the philosophy of psychology and is the author of The Bounds of Cognition (2008).}
\end{adjustwidth}

Kenneth Aizawa cautioned attendees against radical embodiment theses that, he argues, threaten key insights of the cognitive revolution. The cognitive revolution refers to the rise of modern cognitive science and the decline of behaviorism as the dominant approach in psychology. Aizawa holds that cognition should be understood as computation over representations bearing non-derived content, that is, content that does not depend on the attributions, interpretations, conventions, etc. of other agents. This view of cognition has a venerable history but is far from uncontroversial, as Aizawa acknowledges.  What is important is that it draws a distinction between cognition and behavior, and lays out one plausible view about about what cognition could be. 

Azaiwa argues that this distinction between cognition and behavior is central to the cognitive revolution and that neglecting it undersells the importance of our internal cognitive processes. For example, a view which holds that cognition is the maintenance of a certain kind of brain-environment system downplays the importance of the internal cognitive machinery that facilitates our interactions with our environment. To make his argument, Aizawa considers the curious phenomenon of neuromuscular blockade, wherein certain drugs inhibit communication at neuromuscular junctions, causing temporary paralysis. Despite this paralysis, patients remain entirely conscious and cognition is not substantially inhibited. 

Hence, it seems possible for cognition to continue in the absence of any significant bodily movement or environmental interaction, and this supports the distinction between cognition and behavior. This isn't to say that brain-body or brain-environment interactions are unimportant in learning, problem solving, or other cognitive tasks. Rather, it is to say that these interactions can be studied precisely as interactions between a cognitive system and its body/environment, where the distinctive contributions of each can be understood and appreciated. Viewing these interactions themselves as cognition confuses these distinctive roles.  

\noindent\textbf{Discussion:} 

One commenter added that patients suffering from locked-in syndrome have similar experiences to those on neuromuscular blocking drugs, and that the former have even written entire books through blinking alone. The commenter argued that locked-in patients clearly developed their mental abilities through past interactions with the world but that their current abilities cannot be explained in these terms. Aizawa concurred, adding that radical embodiment theorists often fail to adequately distinguish between cognition itself and the causal contributors to cognition and cognitive development. Taking a more critical perspective, Lisa Miracchi and another commenter argued that the absence of brain-world interactions in these exceptional cases does not undermine the constitutive relevance of external factors (e.g., to the ascription of efficacious mental content). Another commenter asked  whether non-embodied systems such as GPT-3 would satisfy Aizawa's definition of cognition. Aizawa argued that cognition is probably best understood as a certain kind of computation over non-derived representations, and that many modern AI systems may be using the wrong algorithms for manipulating their internal representations. The right algorithms are hard (at present) to identify, but we can gain some insight by comparing how humans and animals solve problems to how AI systems solve problems.

\subsection[``The Ad Hoc Construction of Meaning'' (Daniel Casasanto)]{``The Ad Hoc Construction of Meaning''}
\begin{adjustwidth}{1cm}{1cm}
\onehalfspacing
\textit{Daniel Casasanto is an Associate Professor of Psychology and Human Development at Cornell University, where he leads the Experience and Cognition Lab.}
\end{adjustwidth}

Daniel Casasanto argued that concepts, categories, and word meanings are constructed in an ad hoc manner during ordinary cognition. More specifically, he holds that ``Meaning is a dynamic pattern of information that is made active transiently, as needed, in response to internally-generated or external cues.'' On this view, the physical, social, linguistic, mnemonic, and physiological circumstances in which we encounter a word prompt us to form a distinctive (and likely unique) pattern of neuro-cognitive activity, which structures our understanding of the word (or idea) and enables further cognition. This flexible and dynamic approach helps to explain how humans can very easily use words in radically new and different ways across a variety of contexts. 

Forming these patterns of activity, on Casasanto's view, requires the recruitment of language areas of the brain as well as motor areas. For example, the same words in different contexts might activate different motor areas. For example, consider pairing the statement ``it's very hot here'' with two images, one of a desert and one of a closed window. In the second case but not the first, the sentence will activate motor areas of the brain since the context suggests that the statement is an indirect request for action. In this way, the nature and meaning of the words is shaped by context, and this phenomenon is reflected in the brain. 

One might argue that such cases are unusual or that there is a core meaning of words that is context-independent in contrast with metaphorical extensions of words that are context-sensitive. In contrast, Casasanto argues that actual usage of words is too flexible to support the idea of core meanings. That isn't to say that the conventional meanings (e.g., the dictionary definitions) of words are irrelevant---far from it. The point is that these meanings are important cues, but not in a way that sets them apart categorically from contextual factors. The conventional meaning of a word is relevant to but not determinant of the many non-conventional ways we might use it. 

\begin{center}
\textit{To watch the entire talk, please visit: \href{https://youtu.be/VG85PD0YiBw}{https://youtu.be/VG85PD0YiBw}}  
\end{center}

\noindent\textbf{Discussion:} 

The discussion covered a diverse range of issues. The first was whether evidence for the relevance of motor regions to word understanding was all correlational. Casasanto clarified that experimental interventions (e.g., using transcranial magnetic stimulation) had shown that impairment of motor regions impairs understanding of relevant words. Several commenters pushed back, emphasizing the importance of conventional meanings, articulating different ways in which conventional meanings play a critical role in understanding and using language. Casasanto largely agreed about their importance, but maintained that this was consistent with views about the relevance of context. Finally, the discussion turned towards the role of motor areas in language learning and whether learning might be enhanced by approaches that engage motor areas. Casasanto agreed with the idea and cited research suggesting that learning could be enhanced in this way. 

\subsection[\hspace{1cm}General Discussion: Day One]{General Discussion:} 

Brian Cantwell Smith (University of Toronto) led the general discussion for the first day. Smith focused his comments on the idea of semantic efficacy and related issues of scientific explanation. Smith suggested that some systems (such as cognitive systems) are only intelligible in terms of things that don't make an immediate causal difference in those systems. For example, the activity of a binary adder circuit in a computer is most intelligible in terms of binary addition. Nevertheless, the fact that we can think of the adder registers as containing zeros and ones does not mean that the circuit itself is sensitive to the meaning we assign to the states of those registers. There is an underlying physical process which governs the behavior of the circuit, and this processes is not sensitive to what we take certain elements of the system to stand for. The task of explaining how the circuit works requires to say what it is about the system that makes the binary adder interpretation unique in its ability to render the behavior of the circuit intelligible. This suggests that the intelligibility of cognitive processes may not depend whether we actually regard the meaning of a mental representation as making a causal difference.

The subsequent discussion revealed not so much disagreement with Smith's framing, but a conflict of alternative conceptions of causation and explanation. One group was happy to vest causal powers in the physical (or at least sub-semantic) elements of cognitive systems, and treat the appearance that content has a causal role as a phenomenon to be explained by cognitive scientists. Another group preferred to attribute causal powers to higher-level things (such as mental representations and their contents) and to appeal these things in giving a causal explanation of the system's behavior. Despite these difference, both views affirm the ability or apparent ability of mental content to make a causal difference and agree that understanding and explaining this is vital to understanding cognition.

\subsection[``Language is a Neuroenhancement'' (Guy Dove)]{``Language is a Neuroenhancement''}
\begin{adjustwidth}{1cm}{1cm}
\onehalfspacing
\textit{Guy Dove is a Professor of Philosophy at the University of Louisville. He specializes in the philosophy of psychology, drawing on insights from anthropology, cognitive science, linguistics, neuroscience, and other related fields.}
\end{adjustwidth}

Guy Dove began his talk with some remarks about the value of philosophy in approaching scientific questions. He argues that philosophy has a valuable role in exploring the relationships between ideas and in developing speculative theses. To borrow an example from another domain, philosophers have played a significant role in laying out possible ways of understanding what moral claims are and how they are to be evaluated. The philosophical accounts have since become hypotheses for psychologists studying meta-ethical reasoning. The hypothesis promoted by Dove is that language is a scaffolding for thought and a form of extended cognition. In particular our experience of learning language and its formal structure provides us with a valuable set of cognitive tools that we can deploy across a range of different tasks. This view he calls ``LENS Theory'' or ``Language is an Embodied Neuroenhancement and Scaffold.'' 

A major problem that this view is meant to address is the ``symbol ungrounding problem.'' The symbol \textit{grounding} problem is the problem of how internal formal symbols (e.g., those realized in a mind or AI program) come to mean or refer to things in the world. Many embodied accounts suggest that it is the connection between these symbols and our sensory-motor system that connects them to the world. However, humans also have a remarkable ability to abstract from the literal meanings of words (in the use of analogy and metaphor) and to use arbitrary symbols as tools for thinking. If the meanings of symbols (including words) are firmly grounded in our sensory-motor systems, what explains our facility with uses of symbols that run roughshod over their normal sensory-motor associations (e.g., reasoning about the actions of an international ``body'' of scientists or the social consequences of a ``family'' of emerging technologies)? This is the symbol ungrounding problem. Learning language with its arbitrary connections between sounds/symbols and meanings, Dove argues, gives us a model for this kind of flexible cognition. Taken together with embodiment, this view seems to do justice to both the grounded nature of cognition and to our facility for abstract reasoning and the flexible deployment of words and concepts (a point stressed by Daniel Cassasanto on the preceding day).  

\begin{center}
\textit{To watch the entire talk, please visit: \href{https://youtu.be/y6TJ6BopFeE}{https://youtu.be/y6TJ6BopFeE}}  
\end{center}

\noindent\textbf{Discussion:} 

The first commenter raised an objection to Dove's hypothesis. He noted that language deficits often leave other forms of cognition remarkably intact. How could this be, the commenter asked, if we draw on our linguistic abilities to support other kinds of cognition? Dove acknowledged that these results present problems for his view, but that other works (beyond that on cognitive deficits) is more supportive of his hypothesis. Another commenter wondered how much the meaning of symbols even mattered in light of larger language models that (with significant if limited success) learn from the syntactic and statistical features of sentences. Dove argued that while he views language and its syntax as a valuable scaffolding for cognition, he also accepts the importance of grounding symbols (especially in our our sensorimotor system). This aspect of his view, Dove suggested, offered something to those concerned about symbol grounding and those impressed by what can be done with syntax alone. More generally, there was some ambiguity in the talk about exactly how the capacity for language relates to our capacity for abstract thinking. In the talk, Dove framed language as a scaffold for certain kinds of abstract thinking. In the discussion, he framed language not as the source of abstract reasoning but as a special form of elastic thinking (a la Cassasanto). This seems compatible with the position, present in the talk and discussion, that language enhances our capacity for abstract thought but is not strictly necessary for it.

\subsection[``Language, Perception and Action Shape the Human Conceptual System'' (Louise Connell)]{``Language, Perception and Action Shape the Human Conceptual System''}
\begin{adjustwidth}{1cm}{1cm}
\onehalfspacing
\textit{Louise Connell is a Professorial Research Fellow at Maynooth University and holds a visiting researcher position at Lancaster University. She is Principal Investigator for a European Research Council grant on the the role of language in complex cognition.}
\end{adjustwidth}

Louise Connell detailed her research on the relationship between word similarity judgments made by humans and those made by machine learning models using different features of words. A major goal of this research is to determine which features of words (e.g., co-occurrence statistics) best predict human similarity judgments. This work is especially important for present purposes since accounts of embodied cognition hold that many of our concepts are grounded in sensorimotor experience. This suggests that associations between words and sensorimotor experience (e.g., ``sandpaper'' with touch or ``rainbow'' with sight) will predict a significant part of humans' similarity judgments. More specifically, it suggests that words with similar sensorimotor associations will be judged to be more similar in general. 

To assess this prediction, Connell gathered data on how closely people associate various words with different body parts and sense modalities. She also considered data on word distribution (e.g., their rates of co-occurrence), word hierarchies (i.e., how words form nested categories), and overlap between features associated with different words (e.g., ``swims'' is associated with both ``fish'' and ``duck''). Models using each type of data were then compared on their ability to predict human similarity judgments. While models that used sensorimotor associations as predictors were consistently successful, these models were never the \textit{best} predictor for human performance on any of the word similarity tasks. The best predictor was always one of the two distributional models, but one of the two distributional models was also among the worst predictors for each data set.     

To gain further insights, Connell extended her research to include other semantic tasks. For example, she considered human performance on category production tasks (i.e., tasks where participants are asked to produce examples of a given category). The sequences of examples produced during these tasks can (in aggregate) tell us a lot about how closely two words are associated. For example, people would be much more likely to say ``dog'' (and to say it earlier) than to say ``ferret'' when asked for examples of pets. Comparing distributional and sensorimotor models, Connell found that the former were more likely to predict the first word produced, but the latter were more likely to predict where in a sequence a word was mostly likely to appear. 

This suggests, she argues, that distributional information might be employed as a convenient heuristic for producing examples, but that when generating less closely associated examples, we engage sensorimotor concept representations. This idea is supported by the improved performance of a model that uses both distributional and sensorimotor similarity. This model searches both distributional similarity space and sensorimotor similarity space, returning the words according to their maximum similarity to the target category. As it happened, the closest word was often found in the distributional data, whereas similar words were somewhat further out in the sensorimotor space. Overall, the picture presented by Connell was one on which both distributional and sensorimotor information play important roles in our semantic judgments. 
\begin{center}
\textit{To watch the entire talk, please visit: \href{https://youtu.be/9XVgEM5UZOo}{https://youtu.be/9XVgEM5UZOo}}  
\end{center}

\noindent\textbf{Discussion:} 

The first commenter challenged the idea that words are understood and compared along a number of general dimensions. Individuals with category specific agnosias can lose their competence with relatively narrow categories (e.g., an inability to recognize human faces). If this is the case, the commenter argued, then this should lead us to doubt that category understanding is supported by a few highly general dimensions since deficits should apply to much broader swaths of conceptional space (e.g., those involving similar modalities, features, or distributional statistics). While there was not time for a detailed discussion of these cases, Connell argued that category-specific agnosias are rarely as cleanly separated from other deficits as the question suggests and that there might in fact be spillover to other categories.  

Other commenters suggested more exploration of which categories are most fundamental to human understanding. For example, some hoped (and Connell agreed) that follow-up research could look at sensorimotor dimensions with finer granularity (e.g., distinguishing different kinds of haptic feedback) or try to identify distinct similarity measures used by humans for different clusters of words. Others built upon Connell's suggestion that distributional statistics are a heuristic for similarity and suggested that further work might uncover more fundamental ways in which the brain represents categories. For example, some suggested that sensorimotor representations might play this role, while others favored causal network representations. Overall, commenters appreciated Connell's work in gathering data about sensorimotor similarity judgments and were eager to see it extended in various ways. 

\subsection[``Multimodal Grounding for Concrete and Abstract Concepts'' (Penny Pexman)]{``Multimodal Grounding for Concrete and Abstract Concepts''}
\begin{adjustwidth}{1cm}{1cm}
\onehalfspacing
\textit{Penny Pexman is a Professor of Psychology at the University of Calgary. Her research seeks to understand how humans derive meaning from language.}
\end{adjustwidth}

Penny Pexman discussed a diverse range of studies designed to assess the the connections between sensorimotor grounding and word understanding.  Embodied cognition theories generally predict that words with greater sensorimotor involvement should be easier to recognize, process, and remember. Pexman quantified sensorimotor involvement as ``Body-Object Interaction'' (hereafter, ``BOI''), asking study participants to rate ``how easily a human body can interact with a word's referent.'' For example, ``ball'' would typically receive a higher BOI rating than ``democracy.'' She then used these ratings to compare high/low BOI words across different tasks and assess the relevance of BOI when compared to other features of words.

For example, Pexman found that high BOI words were more easily recognized than low BOI words---even when those words were otherwise matched along other relevant dimensions (e.g., age of acquisition). In another tasks, participants were asked to assign words to different categories. High BOI words were assigned to categories more quickly and more accurately, and FMRI results confirmed higher motor area activity when classifying these words. In a third task, participants were asked to identify words as either \textit{concrete} or \textit{abstract}, and the words were compared for BOI, imageability, and emotional salience. Both imageability and BOI made this task easier for both categories, while emotional salience made it more difficult for concrete (but not abstract) words. 

Pexman also conducted factor and network analyses of word ratings along a number of dimensions, including a number of sensorimotor ratings more fine-grained than BOI (see e.g., those discussed in Louise Connell's talk). The factor analysis identified six factors. Three of the factors heavily weighted sensorimotor dimensions. The first emphasized socialness along with dimensions such as audition, mouth involvement, and head involvement. The second emphasized foot/leg and torso involvement heavily. The third emphasized hand/arm involvement along with vision and touch. The other three factors weighted distributional information, emotion, and lexical properties, respectively. Despite the prevalence of factors weighting sensorimotor dimensions, the network analysis revealed that a word's association (or lack thereof) with interoception (i.e., the ability to discern one's internal states such such hunger or fatigue) was the most central of the dimensions. In the factor analysis, interoception was most closely associated with emotional dimensions as opposed to bodily dimensions (such as head or torso involvement). Taken together, Pexman argued that her results support a multi-modal view of how concepts are grounded, where the most important dimensions in each case are shaped by the task demands.    

\begin{center}
\textit{To watch the entire talk, please visit: \href{https://youtu.be/23XEannltlY}{https://youtu.be/23XEannltlY}}  
\end{center}

\noindent\textbf{Discussion:} 

The first commenter asked Pexman to clarify the prospects for views that give embodiment pride of place in symbol grounding. Pexman clarified that she does not believe her work supports strong embodiment positions but does support multi-modal views that take seriously the sensorimotor contribution to grounding. The next commenter asked whether abstract concept processing benefits from sensorimotor processing via imagination or simulation. For example, the idea of ``boating'' as a hobby or sport is fairly abstract, but it certainly involves many motor activities that individuals might associate with it (e.g., raising a sail, battening down the hatches). In response, Pexman noted her work on people playing the game Taboo (where players attempt to communicate a particular word without using the word itself or a handful of closely-related words). Players often used easily imaginable and highly-concrete metaphors to communicate abstract concepts (e.g., using ``dog'' to communicate ``loyal'') that appeal to our imagination. It would not be surprising if humans reached for similarly concrete and imaginable associations when processing abstract concepts in other contexts. Another commenter questioned the empirical basis for the idea that the brain simulates objects and situations in order to reason about relevant concepts, noting that the direction of causation can often be hard to establish in psychological research. For example, do we simulate a dog to help us reason about loyalty or does reasoning about loyalty simply call to our imagination salient examples of loyalty (such as dogs)? The ensuing discussion suggested deeper disagreements about exactly what would constitute evidence for simulations in the brain and, more generally, for the idea that sensory motor areas play a causal role in reasoning about concepts.

\subsection[\hspace{1cm}General Discussion: Day Two]{General Discussion:}

Andrew Lampinen (Deep Mind) began the general discussion from his own work on reinforcement learning agents. He noted that, as Guy Dove suggested, language can act as a kind of scaffolding in reinforcement learning, and that adding linguistic data to training improves performance and results in better abstraction--even in tasks where language is not directly relevant. He also suggested that concreteness might be better understood in relative terms. For example, what might at first be an abstract concept in logic or mathematics (e.g., addition) might later be used as a concrete example of a still more abstract concept (e.g., commutativity). This seems to be an important feature of how we learn and how we build on prior understanding.  Lampinen also noted that it seems important to consider and distinguish situatedness and embodiment in order to understand the contributions of environment and body, respectively. Finally, he asked what participants thought couldn't be learned from examples of language usage alone (e.g., in large corpora of text) and what implications embodiment has for the actual implementation of AI systems. 

The open discussion began with a request for Lampinen's own opinion about what can and can't be learned from large language corpora. Lampinen was unsure, but noted the often surprising pattern of success and failure in large language models: these systems often fail in some seemingly easy tasks while succeeding in some seemingly difficult tasks. Another comment argued that the mark of generalization is the ability to extend knowledge to unusual and unlikely situations and so statistical learning (by itself) faces fundamental limits to its generalization abilities. Lampinen responded that some kind of symbol system that can learn generalizable rules would likely augment the abilities of existing systems. 

Other commenters (responding to Lampinen's question about AI implementation) pressed the idea that embodying existing systems might improve their performance and, especially, their robustness and generalization abilities. Lampinen conjectured that even though human children encounter fewer words during the course of language learning, they might actually get more bits of data due to their situatedness and embodiment. Words are often encountered in particular sensorimotor situations that contextualize them in important ways. That said, he was unsure whether this was more a matter of getting richer and more realistic data than of embodiment and situatedness, per se. Further, some commenters were skeptical that embodiment or situatedness would be of much value to AI systems without substantial architectural changes that enhance abstract reasoning and allow these systems to leverage their experience in the world.

\subsection[``Self-Generated Learning May Be More Radical Than First Appears'' (Linda Smith)]{``Why ``Self-Generated Learning'' May Be More Radical and Consequential Than First Appears''}
\begin{adjustwidth}{1cm}{1cm}
\onehalfspacing
\textit{Linda Smith is a Distinguished Professor and Chancellor's Professor of Psychological and Brain Sciences at Indiana University, Bloomington. She is the Principal Investigator of the Cognitive Development Lab.}
\end{adjustwidth}

Linda Smith discussed her work on how babies learn and argued for the importance of self-generated learning. A central principle of Smith's work is that animal brains undergo experiences, learn from those experiences, and produce behaviors that shape their experiences in the future. This feedback loop allows animals to produce behaviors that enhance their opportunities for learning. In addition to this principle, Smith argued for three key theses in her talk: (1) Bodily development determines what data infants have access to, (2) behavior creates data for learning in real-time, and (3) the systems of knowledge are shaped by the constraints of space, time, and context. 

To support these theses, Smith outlined her work documenting what babies experience and when. This research is accomplished by attaching sensors to babies and recording what they see and hear---both in ordinary life and in experiments. As babies develop physically, they are increasingly able to control their head orientation and move around their environment. For example, early in life (1-3 months), vision is still developing and babies can mostly shape their experience by moving their eyes to look at nearby objects. Unsurprisingly, human faces (mostly their parents' faces) dominate their visual experience. Later (8-10 months), babies can crawl, but crawling posture keeps their heads down and they mostly see objects on or near the floor. Later still (12-18 months), babies are beginning to walk and they perceive much richer scenes guided by their own exploration. 

These developmental changes shape the opportunities that babies have to learn in real time. For example, as babies develop they see fewer faces and more hands. Often these hands are their own (or perhaps their parents') manipulating objects. Smith recorded the experiences of young children playing with unfamiliar objects alongside their parents. She then tested the children to see which object names they had learned and how those objects had featured in their experiences. Naturally, objects that were close by and centered in their field of view when named were learned more reliably. In addition, other factors such head stability, head-eye alignment, and whether the child held the object also contributed to success. Each of these factors is enhanced by the developmental changes noted earlier (e.g., by having control of one's head, by having one's hands free to grasp, etc.). This is further evidenced by the fact that babies manipulate objects in informative ways as they mature (e.g., by orienting objects along major axes of elongation). 

This tendency to orient objects in informative ways is not entirely surprising given the kinds of physical constraints people have (e.g., our bodies have a preferred up-down orientation due to gravity as well as numerous symmetries). Further, babies exist in a social context with parents who, Smith's research suggests, follow their baby's lead in exploration and also guide that exploration. As a result, there is a close correlation between the objects that parents and babies interact with during play. More generally, exploratory behaviors tend to produce statistical distributions over object interactions similar to those that result from many other processes (e.g., Zipf's Law). Taken together, these phenomena create stable patterns of experience that enhance memory and learning. 
\begin{center}
\textit{To watch the entire talk, please visit: \href{https://youtu.be/U87dqzUO694}{https://youtu.be/U87dqzUO694}}  
\end{center}
    
\noindent\textbf{Discussion:} 

The first commenter asked what would happen if (instead of data sets of random examples) we gave AI systems the data collected from infant behavior. Smith noted that this had been done and that this data did enhance learning. However, she also noted that the ability to learn from this kind of data is not the same as a domain-general learning system that leverages its own behavior to enhance learning across contexts (i.e., by generating useful data particular to that context). Another commenter asked to what extent the independent contributions of different factors could be assessed (e.g., head eye alignment vs. object manipulation). Smith responded that pulling these apart would be difficult and dubiously informative given the deep dependencies between them. The next commenter asked about developmental disabilities that impair the kind of behavior she described. Smith argued that multiple paths to learning exists (e.g., replacing sight with touch in blind children), but that sometimes external interventions are required. For example, physical therapy can help improve muscle tone in children with Down Syndrome and that meeting motor milestones is highly correlated with reaching cognitive milestones. Smith emphasized this result as an example of a practical application of embodied thinking about cognitive development, and she noted that great strides had been made in helping children with Down syndrome reach cognitive milestones that they rarely reached in the past. 

\subsection[``Language Comprehension Requires Affordances'' (Arthur Glenberg \& Cameron Jones)]{``Language Comprehension Requires Affordances''}
\begin{adjustwidth}{1cm}{1cm}
\onehalfspacing
\textit{Arthur Glenberg is a Professor Emeritus at Arizona State University, where he leads the Laboratory for Embodied Cognition in the Department of Psychology. Cameron Jones is a PhD student in Cognitive Science at the University of California, San Diego. His research focuses on the role of world knowledge in language understanding and on the impact of language on categorization.}
\end{adjustwidth}

Arthur Glenberg and Cameron Jones presented their recent work on probing large language models for understanding of affordances. To understand the affordances of something is to understand how one might interact with it. Glenberg and Jones's research follows up on Glenberg's earlier work on latent semantic analysis and affordances, applying a similar paradigm to today's language models. In that earlier work, latent semantic analysis models struggled to distinguish objects whose affordances were appropriate in a given context from objects whose affordances were not appropriate. For example, they might ask whether a model regards ``Alice used a map to fan the flames'' as more likely than ``Alice used a rock to fan the flames.'' Glenberg hypothesizes that humans understand many affordances through sensorimotor simulations of the relevant contexts and objects and that language models struggle because they rely on distributional information alone. 

In their recent work, Glenberg and Jones test three language models---BERT, RoBERTa, and GPT-3---using a variety of metrics appropriate to each model. The purpose of these metrics was to assess whether each model can discriminate between appropriate and inappropriate uses of words, where the referents' affordances are the salient difference between each usage. For example, consider the sentence, ``Tom was tired of blowing on the coals, so he grabbed a rock/newspaper to fan the flames.'' Humans can immediately appreciate that rocks make poor fans. 

Glenberg and Jones found that all three language models struggled with these tasks. GPT-3 performed the best, but it clearly lacked some information or understanding that the human subjects possessed. Glenberg and Jones hope to follow up on this research with further probes that promise to reveal how GPT-3 succeeds (insofar as it does) and how humans approach these problems. 
\begin{center}
\textit{To watch the entire talk, please visit: \href{https://youtu.be/Xn2JKNGYans}{https://youtu.be/Xn2JKNGYans}}  
\end{center}

\noindent\textbf{Discussion:} 

The first commenter wondered whether we already had sufficient reasons to say that large language models lack understanding---setting aside the speakers' research. After all, if we know that large language models learn distributional information about language, then doesn't this suffice to show that these models do not learn affordances? Another commenter disagreed, noting that there was considerable debate within AI about whether an existing AI system can be said to understand things that go beyond strictly distributional information. 

Jones pointed out that building a case for the deficits of current approach must begin with documenting differences in behavior between humans and AI systems, differences that later research goes on to explain. Linda Smith suggested that future work might go beyond the kinds of training data ordinarily used for training these models (large corpora of written texts) and use the kind of language parents use with children. This, she argued, might give us a better sense of what can be learned from language---which she noted was especially helpful for children with disabilities (e.g., blindness) and could substantially compensate for their deficits. 

Another concern was that perhaps affordances are not as fundamental as causal knowledge. After all, we can make causal inferences about things with which we could never interact. For example, we can infer that if the moon fell into the sun, it would be melted and ultimately vaporized. Glenberg suggested that we might reason about these unusual cases by analogy. After all, we know (for example) what happens when we put ice into a hot drink or forget a pot of boiling water. 

Finally, there were some concerns that the earlier research by Glenberg (and thus the examples used in the current study) had been included in the training data for GPT-3. Jones said that there was no way to know this, but that AI performance on other tests had been shown to suffer when examples of those tests were removed from training data. This is likely to be an issue for a number of different measures of AI intelligence as more and more publicly available data is subsumed into training material.

\subsection[``System 2 GFlowNets'' (Yoshua Bengio)]{``System 2 GFlowNets: Inductive Biases Towards Machines that Understand and Think''}
\begin{adjustwidth}{1cm}{1cm}
\onehalfspacing
\textit{Yoshua Bengio is a Professor of Computer Science at Université de Montréal. He is the Founder \& Scientific Director of the Mila AI Institute in Quebec, and he co-directs the CIFAR Learning in Machines \& Brains program and acts as Scientific Director of IVADO.}
\end{adjustwidth}

Yoshua Bengio laid out his team's recent work on GFlowNets---a novel deep learning architecture. Setting aside technical details, he focused on what role this  architecture might play in advancing machine intelligence.\footnote{Bengio lays out technical details in a helpful tutorial found here: \url{https://milayb.notion.site/GFlowNet-Tutorial-919dcf0a0f0c4e978916a2f509938b00}.} Bengio argues that existing AI systems are most similar to ``System 1'' thinking in humans. They excel at perception and at learning quick and dirty heuristics for solving problems. What these systems mainly lack are mechanisms for engaging in the slower, sequential, and rule-based reasoning that characterizes ``System 2'' (Kahneman, 2011; Evans \& Stanovich, 2013). System 2, Bengio suggests, is primarily responsible for the ability of humans to generalize their knowledge to a wide range of domains and to acquire new knowledge without the extensive training required by deep neural networks. 

Bengio, following Dehaene (2014) and others, holds to a global workspace view of higher-level reasoning in humans. On this model, a number of more specialized modules compete for limited opportunities to broadcast information in a shared workspace. These limitations acts a kind of information bottleneck, which filters the information that will be globally broadcast and requires modules to distill relevant information from their inputs. Because different modules can alternatively take the mic, so to speak, this also pushes cognition toward sequential processing. Further, individual modules must encode their outputs in a way that allows other modules to utilize these outputs. The result is that cognition combines thoughts into a task-appropriate sequence for processing incoming information. This combinatorial process, Bengio argues, allows generalization via the mixing and matching of different thoughts into (potentially) entirely novel sequences. 

GFlowNets take inspiration from Bengio's view of human cognitive architecture. GFlowNets also contain many parts whose outputs can only be passed to future processing through an information bottleneck. More specifically, at each time step, different parts of the network compete to produce an activation state that becomes parts of the network's input at the next time step. We can interpret this network as learning the transition probabilities between states and model this information as a graph of states with edges representing transition probabilities. How these states link up to the world will vary from application to application, but states are ultimately assigned values which are used to update these probabilities and (via training) produces more and more valuable sequences. This allows different parts of the network to specialize in producing appropriate outputs in different contexts and allows the network to generalize to novel situations by changing the sequence of states it produces. Early results with this architecture are promising, but for present purposes, the key results are (i) the identification of a gap in existing deep learning architectures (i.e., the absence of System 2 reasoning) and (ii) an implementable hypothesis about how this gap might be filled.\begin{center}
\textit{To watch the entire talk, please visit: \href{https://youtu.be/Q1fw75InQZE}{https://youtu.be/Q1fw75InQZE}}  
\end{center}

\noindent\textbf{Discussion:} 

The discussion focused on clarifying details of the GFlowNet architecture and the ideas about cognition that motivate it. For example, a neuroscientist asked whether Bengio thought that the neocortex played an important role in enabling human intelligence and whether his approach had any analog of it. Bengio argued that the neocortex is important, but as yet, not part of his approach. He suggested that the neocortex has a role in top-down attentional processes and that the global workspace/information bottleneck is about bottom-up competition. GFlowNets focus on realizing the latter. The next commenter asked whether neural networks are actually good at System 1 thinking given their problems with brittleness, forgetting, limited generalization, etc. Bengio suggested that many of these problems might be resolved with a better, higher-level system for leveraging their strengths. If, for example, cognition labor can be allocated via the competition of more specialized systems and if their outputs can be combined into novel sequences for novel circumstances, this might mitigate the problems with existing architectures since a single system will not (by itself) need to generalize to the full range of circumstances.  

Others worried that Bengio's approach, while it tipped its hat to multi-modal grounding in lower-level processes, didn't take embodiment seriously enough. Could a system, for example, effectively learn causal relations in the world without undertaking experimental interventions and observing their effects? Others worried that Bengio had focused too much on the role of epistemic success (i.e., correctly modeling the world) rather than practical success (i.e., an agent achieving its goals). That said, whether these concerns prove to be warranted will depend to a large extent on how GFlowNets are applied. In a reinforcement learning agent embodied in a virtual environment trying to achieve some goal, both concerns would be addressed, at least to some degree. In a non-embodied NLP system learning to produce probable strings of words from a text corpus, they would not be.

\subsection[\hspace{1cm}General Discussion: Day Three]{General Discussion:}

Tyler Marghetis (UC Merced) focused on two key points in setting the stage for the group discussion. First, he argued for the importance of learning on different timescales. For example, there is learning on evolutionary time scales which prepared brains to learn from their environments. There are also developmental timescales, in which learning occurs as the body and mind develop, and, finally, there are short timescales where we acquire particular knowledge or skills. Learning at each of these timescales is shaped in different ways by our embodiment and situatedness. Even on the shortest timescales, events in our environment often unfold sequentially, making the kind of sequence learning proposed by Bengio particularly apt. Second, Marghetis also asked for clarification on what exactly it was about about embodiment that mattered. In one sense, there might be an active role for the body in cognition. In another sense, embodiment (and associated developmental phenomena) might structure learning in helpful ways. For example, children (as discussed by Linda Smith) progress through an increasingly rich visual world as their bodies and visual system develop. This requires that babies ``start small'' and gradually build up their knowledge of the world.

Melanie Moses began the open discussion by asking whether the structure of the brain, with its complex inter-level dependencies, is a major impediment to our understanding of how it works. Linda Smith replied that in the (similarly complex) context of embryonic development, it was hard to pin down specific causes of developmental changes but that scientists could identify a host of ``permissive conditions'' that facilitate important changes. This suggests that we could make progress in understanding psychological development and learning despite the complex organization of the brain. Melanie Mitchell followed up with a question for Smith, asking whether she thought that predictive coding models of learning were on the right track. Smith responded that it is likely only correct at a fairly high level of abstraction and that more fine-grained accounts of learning will be required. Further, she noted that when other accounts have championed a single idea about what motivates learning, they have proven too simplistic. For example, curiosity is very helpful in motivating learning, but it also appears to be the case that children's returning to familiar objects and behaviors is an important part of solidifying learning and avoiding forgetting. 

Another commenter asked whether combinational and sequential circuits might be useful metaphors for different modes of cognition and learning, for example, for the differences between the kind of fast, heuristic, and highly-parallel processing associated with ``System 1'' and the slower sequential processing associated with ``System 2''. Another commenter suggested that these and other models might be useful metaphors for cognitive processes due, in no small part, to their being implementable in software. As such, they can easily be studied, experimented on, and changed in silico---even if they prove to be wrong in several details when compared to models that are not so easily studied. Taking a contrary position, one commenter suggested that these metaphors are often too tempting for their own good and invite us to ignore important differences between machine learning models and what happens in the brain.

\subsection[``Embodiment Without Situatedness'' (Nick Cheney)]{``Embodiment Without Situatedness: A Case for an Embodied Perspective on Neural Architecture Search''}
\begin{adjustwidth}{1cm}{1cm}
\onehalfspacing
\textit{Nick Cheney is an Assistant Professor of Computer Science at the University of Vermont. He directs the UVM Neurorobotics Lab and is affiliated with the Vermont Complex Systems Center.}
\end{adjustwidth}

Nick Cheney outlined his recent work on optimizing bodies and bodily control systems in virtual environments. This work considers alternative optimization strategies, with a special focus on discovering better body plans in the presence of fragile body-control system coupling. On Cheney's view, ``embodiment'' concerns the physical properties of an agent's body, and ``situatedness'' concerns the role of an agent's ongoing interactions with its environment. As an example of the former, Cheney noted that designing ``soft'' robots whose bodies flex and deform in various ways can often solve problems that might otherwise require careful planning or control. For example, a hand with rigid parts will not naturally conform to the shape of a held object and a control policy must be designed to ensure that hands assume the right shape to maintain grip. A hand with soft or flexible parts will naturally conform to the surface of grasped objects if designed correctly. 

Cheney next outlined three key insights from his research in this area. First, he argues that network topology alone is a powerful, underutilized, and body-like feature of cognitive systems. Second, he argues that neural architecture search has generally handled brain-body coupling poorly, and that brains and bodies cannot be optimized in isolation. Finally, he argues that body and control system optimization should occur on different timescales. 

Cheney then outlined a common problem in evolving body designs using genetic algorithms: a few basic body designs emerge early and few arise in later generations. Further, evolving a control system (e.g., a neural network) alongside these body plans doesn't seem to result in new or more diverse designs. Instead, Cheney found that if one alternates between long and slow periods of body plan evolution and short and fast periods of control system evolution, then new and better designs do emerge. The idea is that the former is akin to learning over evolutionary time, and the latter is akin to learning over developmental time. Learning over developmental time allows control systems to adapt to changes in body plans and thereby allows those new plans to be quickly put to use. Without this rapid evolution, new designs will often suffer a fitness disadvantage until control systems adapt to the new plan. All this suggests that learning is improved when it takes place over multiple timescales, with the body adapting slowly and a control system adapting quickly.

\begin{center}
\textit{To watch the entire talk, please visit: \href{https://youtu.be/C-R63mN9BC0}{https://youtu.be/C-R63mN9BC0}}  
\end{center}

\noindent\textbf{Discussion:}

The first commenter, Arthur Glenberg, wondered what implications Cheney's research had for cognitive psychology given that Cheney focused on relatively simple virtual organisms with correspondingly simple control systems. Cheney wasn't sure, but suggested looking for an analogy to learning and optimization at higher levels, and Glenberg added that the work may point to realistic constraints (e.g., arising from brain-body coupling) as a way to enhance learning. The next commenter asked whether sub-networks pulled from randomly initialized networks could be improved by training. Cheney responded that they could, but that the point of the experiment was mainly to show what could be done by selecting the right network architecture without optimizing weights. Cheney also added that choosing among the existing sub-networks often led to solutions with comparatively better generalization to unseen data (likely by avoiding overfitting). Another commenter asked about the genotype to phenotype mappings in neural network evolution and what properties of these mappings enhanced optimization. Cheney noted that this was a complicated subject but that evolutionary algorithms seem to work best when artificial genes are related to the neural networks they describe via a naturalistic process. For example, these genes might shape how a network is built rather instead of specifying each weight or connection---much like genes in nature primarily influence natural neural networks by shaping the course of brain development rather than specifying particular neurons or their connections. 

Melanie Mitchell noted that much of the evolutionary process and the resulting designs might not be intelligible without reference to processes at multiple levels and multiple time scales. She then asked what idea of ``intelligence'' Cheney was employing when he attributed intelligent behavior to the creatures that evolved in his experiments. He replied that defining intelligence is a difficult and highly philosophical questions, but that simple animal behaviors were a tractable starting point on the continuum of animal intelligence. Another commenter then added, responding to Mitchell's first remarks, that cognitive science needs both diachronic (over time) explanation and synchronic (at a single time) explanation for our cognitive abilities. This was meant to suggest that while there may be very interesting evolutionary or developmental explanations of evolved designs, there remains an important question about how to explain the behavior of a system here and now. This continued a larger theme of the workshop---determining exactly how developmental history should figure in our explanations of cognition.

\subsection[``Are Foundation Models Castles in the Air?'' (Percy Liang)]{``Are Foundation Models Castles in the Air?''}
\begin{adjustwidth}{1cm}{1cm}
\onehalfspacing
\textit{Percy Liang is an Associate Professor of Computer Science and Statistics at Stanford University. His research seeks to develop trustworthy systems that can communicate effectively with people and improve over time through interaction.}
\end{adjustwidth}

In his talk, Percy Liang explored the prospects and limits of foundation models. Foundation models (e.g., large language models) are deep neural networks pre-trained on large, broad data sets and then adapted to specific tasks. The foundation model paradigm might be distinguished from the transfer learning paradigm in that foundation models are not trained on one specific task and then asked to complete another. They are trained on a highly-general task (e.g., next word prediction) and then asked to complete more specific tasks (e.g., summarizing text). 

This generality and the nature of the training data (e.g., large language corpora) raise concerns about whether the current limitations of foundation models can be overcome by merely scaling up this approach. To be fair, large language models have especially benefited from huge increases in parameter counts and data set size. At the same time, they also exhibit a number of serious failure modes. For example, they are vulnerable to data poisoning attacks, in which an adversary manipulates the behavior of a model by seeding its training set with examples of the desired behavior (e.g., a large number of documents with false or misleading information). Foundation models are also found to replicate ethically questionable biases in their training data (e.g., replicating racial or gender stereotypes). Many scholars doubt that relatively passive learning aimed at reproducing the statistics of natural data (e.g., producing probable strings of words) can lead to human-like intelligence.

Despite these concerns, the surprisingly successes of these systems, argues Liang, should motivate a humble assessment of their abilities that recognizes our limitations in imagining their future abilities. As part of this assessment, Liang considers the differences between foundation models and the human mind, and he aims to study how these differences might affect performance. For example, Liang notes that, unlike foundation models, humans learn from multi-modal data, that our data involve actions and their consequences, that our own activity can generate new data, and that humans have a more structured and more modular neural architecture. Future research should attempt to discover what is lost by omitting these features and which would be best incorporate into foundation models. For example, attempts have already been made to use images combined with text to train models that can generate images from captions. 

Overall, Liang argues that we have reason to be optimistic. For example, he notes that blind children often learn to use color terms with surprising success despite limited color experience, and language contains many examples of color ascriptions and associations from which they can draw. This is analogous, Liang suggests, to an unsupervised clustering algorithm identifying distinct categories to which things belong but not having specific labels (analogous to color experiences) to associate with them. He also notes that adding action labels to image sequences depicting the steps of various processes (e.g., making a peanut butter sandwich) helps foundation models learn to predict such sequences, but that this mostly helps to shorten training time rather than to improve performance. As Liang argues, this suggests that there is much more to infer from large data sets than we might intuitively believe, and thus more promise in further scaling up foundation models. 

\begin{center}
\textit{To watch the entire talk, please visit:  \href{https://youtu.be/oD9d\_MZV5Og}{https://youtu.be/oD9d\_MZV5Og}}  
\end{center}

\noindent\textbf{Discussion:} 

The first commenter argued that just because the foundation model approach could work \textit{in principle}, doesn't mean it is practically feasible---much less the best approach on offer. Foundations models, after all, are ballooning in size, and by better understanding intelligence, we might well find simpler and more practical models. While acknowledging the commenter's general point, Liang argued that scaling up these systems might still result in large increases in performance before the practical limits of scaling are reached. Other commenters were worried that not enough had been done to explore and articulate the weaknesses of foundational models (both in the talk and in wider research). Liang responded that he had conducted a great deal of research on their limitations and conceded that these limitations were numerous. At the same time, he noted that many limitations had been overcome by using larger models and larger and more diverse data sets. This is why, he argued, it is important to consider whether there is any in principle reason to think this progress will not continue. However, it was not totally clear exactly what kinds of changes to foundation models would count as mere scaling up. For example, incorporating multi-modal data seems like a difference in kind, and Liang was explicitly not committed to any particular neural network architecture (e.g., transformers) or relationship to the external world, allowing that foundation model might be used as part of a robot's control system. If significant changes of this kind are required, that seems like a vindication of those emphasizing the importance of embodiment, multi-modal grounding, and so on.

\subsection[``Could a Self-Supervised Foundation Model Understand Language?'' (Chris Potts)]{``Could a Purely Self-Supervised Foundation Model Achieve Grounded Language Understanding?''}
\begin{adjustwidth}{1cm}{1cm}
\onehalfspacing
\textit{Chris Potts is a Professor of Linguistics and Computer Science at Stanford University. He currently serves as the Chair of the the Department of Linguistics.}
\end{adjustwidth}

Chris Potts discussed foundation models, continuing on the subject raised in the previous talk by Percy Liang. Potts began by answering his own question in the affirmative, adding that he sees no reason why a foundation model could not learn to understand language. He began by defending this position on intuitive grounds, noting that our intuitions about what systems are capable of understanding language (e.g., in the Chinese Room and Chinese Gym thought experiments) are influenced by the size, speed, and complexity of relevant system. Potts continued by elaborating his own understanding of foundation models, emphasizing their size, generality, and self-supervised training. He also focused on models that learn from sequences of data (e.g., sequences of images, words, events, etc). 

For comparison, Potts considered two cases. First, a model is trained on a large data set of chess sequences---the model is given no explicit signals about what the rules of chess are, what moves are possible, or what constitutes success. Second, another model is given a structured space of actions, many example games, many practice games against itself, explicit reward signals, etc. Potts then asked the audience to consider what the first AI might learn and how it might compare to the second. Potts argued that a great deal could be likely be learned from sequences alone, but that some minimal reward signal might be required. 

Potts then addressed how we might test AI systems for understanding, discussing two complementary methods. The first and more traditional method focuses on performance benchmarks designed to test specific competences. The other is more explicitly cognitive in nature and involves probing AI systems for causal models of relevant phenomena. For example, Potts' research identified activations associated with the presence or absence of distinct (but causally related) stimuli. He then tested whether causal models of the interactions between these activations mirrored a causal model of the corresponding stimuli, finding that in several important cases (e.g., BERT language models) they did. Potts has conducted similar research which suggests that large language models have models of entailment relations between concepts. Overall, Potts took an optimistic view of what such tests might tell us about future AI systems, while remaining realistic about the limitations of current AI systems. 

\begin{center}
\textit{To watch the entire talk, please visit: \href{https://youtu.be/BJ-8t8xrxq8}{https://youtu.be/BJ-8t8xrxq8}}  
\end{center}

\noindent\textbf{Discussion:} 

The first commenter questioned whether, when larger models use bigger data sets from more modalities, the hardware limitations will prevent us from reaching the full potential of foundation models. Further, the commenter worried that some kinds of data (e.g., about one's own emotional responses) might be readily available for human beings but very hard or impossible to acquire for AI training. Potts argued that existing models are quite limited relative to humans---both in their complexity and in the data they receive during training. He noted that efforts are underway to collect large multi-modal data sets based on infant experience and train foundation models on this data. Another commenter worried about the need for active learning (drawing on the central theme of Linda Smith's talk). Potts again noted the possibility that some light feedback from humans (e.g., about chess performance) might be necessary and that other interactions (e.g., conversations with humans based on text it generated) might constitute a form of active learning as the AI could observe and learn from text generated in response to its own production. 

Lisa Miracchi was concerned about the internalist assumptions of Potts's testing paradigm--in particular, she seemed to doubt whether we can really understand the activations in neural networks without considering how those networks are embedded in and interact with their external environment.  Potts was somewhat skeptical of this view and wondered what exactly would be missing in a foundation model given rich multi-model data. Miracchi argued that many (for example) human perceptual processes are not intelligible in terms of a clear distinction between sensory inputs and behavioral outputs. Rather, they are better understood as a kind of coupling with the environment in an ongoing feedback loop of perception and action that makes perception what it is.

\subsection[\hspace{1cm}General Discussion: Day Four]{General Discussion:}

The final day's discussion concerned general thoughts about the workshop as a whole. We draw on this discussion in outlining the major themes in the section that follows. 

\section{Conclusion}

A number of common themes emerged during the workshop, acting as focal points of discussion and debate for a diverse group of attendees. While the discussion was extremely constructive, it also revealed strong differences in opinion, methodology, and aims among the attendees, signalling the need for further research and greater interdisciplinary engagement in this area. Three broad themes arose repeatedly throughout the workshop. First, there was general agreement that embodiment and situatedness play a large role in human development, but participants disagreed about what this implies about learning in AI systems. Second, there was some controversy about the degree to which embodiment and situatedness play an active role in ongoing cognitive processes (especially with respect to sensorimotor grounding and extended cognition). That said, none of the participants denied that embodiment and situatedness play a role in at least some ongoing cognitive processes. Third, there was considerable controversy about whether distal or historical factors arising from the interplay of body and environment play an important explanatory (or perhaps constitutive role) in ongoing cognition. 

In addition to these broader points of discussion, there were also several themes that concerned more specific subjects---especially learning in AI systems. For example, there was the idea that learning does (and should) take place over multiple time scales (e.g., evolutionary and developmental). There was also the idea that large neural networks lack critical elements of human learning, and that large language models in particular often exhibit ``truthy'' behavior (i.e., producing plausible-looking text with little regard for truth or overall consistency). Finally, there was the issue of whether some form of reward or motivation is essential for agents to learn. We discuss each of these themes in the subsections that follow and suggest directions for future research based upon them.

\subsection{Embodiment, Situatedness, and Human Development}

The most important (and least controversial) idea from the workshop was that embodiment and situatedness play an important role in human development. Even among those most skeptical of more radical accounts of embodied cognition, all were quite willing to endorse some version of this idea. Linda Smith presented this idea most forcefully in her talk on ``self-generated learning,'' but several other speakers and commenters also remarked on its importance. Self-generated learning occurs when learners shape the data from which they learn. Smith's talk, for example, emphasized the importance of physical interaction and exploration in learning and development. Children move around their environment, pick up, examine, and manipulate objects, track their parents' gaze, etc. Further, different stages of physical development direct an infant's gaze and behavior, structuring learning in helpful ways. All of these things shape (and ultimately enrich) the data children have about the world. 

One of the more controversial points here was whether what matters in self-generated learning is fully captured in the data gathered by infants about the world and how much depends on self-generation, as such. For example, several attendees suggested taking data acquired by monitoring what infants see and hear and giving it to AI systems as training data. These AI systems (crucially) would not interact with the world themselves, but rather would learn from observing the data generated by infants' interactions with the world. Some attendees were optimistic about the prospect for learning in this way. Others (e.g., Smith) felt that this might be effective, but that a system that learned in this way might not have the kind of generalizable capacity for learning that infants have. Arguably, it is a major weakness to be constrained to learning from data that is carefully collected by researchers and generated by more competent learners. Finally, some were pessimistic that this would work and doubted that learning from data generated by others could ever be as effective as true self-generated learning.  For example, drawing on interventionist theories of causation (e.g., Pearl, 2000; Woodward, 2003), some argued that intervening on one's environment was necessary for testing and refining one's hypotheses about the world's causal structure. In any case, this seems like a vital area for future research and one where important insights might be gained in the relatively near term.

\subsection{Embodiment, Situatedness, and Ongoing Cognition}

While it is difficult to deny that having a body and being situated in an environment facilitates learning (especially during early development), there was disagreement on how and when these factors are relevant to ongoing cognition. To illustrate a case where these factors do seem highly relevant, Clark and Chalmers (1998) consider the benefits of rotating pieces on the screen when playing Tetris compared to rotating them only in the mind's eye. Very few people would be confident enough in their mental rotation abilities to play without rotating the blocks on the screen. Their argument is that (along with the player's brain) the machine, screen, and software are playing an active role in the cognitive task of playing Tetris. A related idea is that physical structures can do work that might otherwise require computation for control. As Nick Cheney discussed, ``soft robots'' use flexible structures (e.g., gripping implements) in ways that would otherwise require careful computer control (e.g., for conforming one's grip to an object). At the workshop, there was little direct pushback against these ideas, but Ken Aizawa argued that studying and understanding cognition requires us to draw a distinction between behavior and cognition. This wasn't to say that interacting with the external world is unhelpful but rather that delimiting the domain of cognition is essential for isolating it as a subject of scientific study. 

Hence, disagreements were not so much about whether interacting with the external environment can help us to perform cognitive tasks, but about whether it is useful for cognitive scientists to draw a distinction between these interactions and cognition proper. Another role for embodiment in ongoing cognition is less direct and figured in many of the workshop's talks. This idea is that logical and syntactic relations between mental symbols are insufficient to connect those symbols to their referents in the external world via perception and behavior. Understanding how the brain succeeds in connecting mental symbols to the external world is the \textit{symbol grounding problem}. There are at least two ways one might frame this problem. 

One is as a general concern about the meaning of mental symbols. On this view, nothing in the brain is sufficient to fix the meaning of a mental symbol. While the brain constrains meaning, the history and environment of an animal also necessary to fix the meaning of its mental symbols. Analogously, the physical properties of one part of a mechanism (e.g., a gear) are not sufficient to fix its function within the mechanism. These properties constrain the functions it might perform, but the part doesn't have any determinate function outside the context of a specific mechanism, since many mechanisms use the same parts for different purposes. Similarly, suppose that in a particular species of frogs, the visual and cognitive systems have evolved to detect flies by tracking dark moving patches in the field of view, representing the presence of a fly with a particular mental symbol.\footnote{Thought experiments involving frog perception have a rich history in this area of cognitive science and were inspired by the empirical research of Lettvin et al.\ (1959).}  Of course, there might be many relevantly different environments in which this approach would be successful (e.g., one in which an ant-eating frog species frequently sees ants crawling on lighter color rocks). Suppose we dangle an ant on a fine string and move it in front of a frog---causing the frog to strike. Whether the frog correctly represented the ant as an ant or incorrectly represented the ant as a fly depends on the meaning of the symbol that the frog uses to indicate a detection. However, as our thought experiment presupposed, nothing in the frog's brain can settle the meaning of the symbol since its visual and cognitive mechanisms are fit for two purposes. We must consider the frog's present and historical ecological niche to determine whether we are dealing with an ant-eating or a fly-eating frog species and assign meaning to the symbol accordingly (Millikan, 1989). 

Another way of thinking about the symbol grounding problem is to view it as a problem of cognitive architecture, and this seems closer to the original framing (Harnad, 1990). Assuming that an important part of cognition involves the manipulation of symbolic representations, we might wonder about how these symbols are associated with things in world (when they are). We might consider the logical or syntactic properties of symbols (e.g., whether one representation can be inferred from another or whether a representation can take an alternative but equivalent form). Suppose we have symbols for ``is a mammal'' ($M$), ``is a dog'' ($D$), ``is a reptile'' ($R$), and ``is a snake'' ($S$). We might infer $Mx$ from $Dx$ or $Rx$ from $Sx$, but given that these categorical inferences are formally identical, what makes $M$ (rather than $R$) count as ``is a mammal'' (rather than ``is a reptile'')? More specifically, how does the brain connect these abstract symbols to things in the world such that we can interact with the world appropriately? One answer is that the way we represent concepts is firmly grounded in our sensorimotor system, and as such, concepts representations come along with sensorimotor associations that allow us to connect abstract reasoning to appropriate environmental interactions. 

Crucially, those concerned with the first version of the problem would not be satisfied with a sensorimotor grounding account of meaning, and would deny that \textit{anything} in the head is sufficient to fix the meaning of many of our mental symbols (e.g., Putnam 1974; Burge 1979). Similarly, those concerned with the second version will not be satisfied by appeals to ecology or evolution---the fact that the frog lives and evolved in a fly-infested rather than ant-infested environment doesn't tell us how the ``fly detected'' symbol is connected to the visual system or to the motor system which actuates the tongue. Returning to the main issue, what might each of these accounts (i.e., sensorimotor grounding and mental content externalism) tell us about the role for embodiment/situatedness in ongoing cognition? In the first case, it seems that embodiment and situatedness are directly relevant since sensorimotor concept representations are taken to be involved in ongoing cognition. However, given that these representations, once acquired, can be used by ``locked-in'' patients, it is unclear whether what matters for ongoing cognition is our embodiment or situatedness, per se. In the second case, much will depend on what external factors one takes to be relevant to meaning. On Lisa Miracchi's view, what matters is often a kind of ongoing coupling to the external environment that looks more like the kind of process involved in the Tetris case. However, this account would seem predict cognitive deficits in locked-in patients for whom this coupling ability is (presumably) significanly impaired. For others (e.g., Millikan, 1989; Dennett, 1990), historical or ecological factors are relevant, and it is difficult to view these as relevant to ongoing cognition. If we fed a pet frog ants, would the fact that wild frogs of the same species eat flies have anything to do with the cognitive process by which the pet frog picks out ants to eat?

\subsection{An Extended Role for Embodiment and Situatedness}

A more controversial idea was whether distal or historical factors related to body and environment play a role in ongoing cognition. For example, while many would affirm that these factors play a role in development and contribute to building an adult cognitive system, fewer would regard them as playing an active role in ongoing cognition. For example, Ken Aizawa asked the audience to consider patients experiencing near total paralysis induced by neuromuscular blocking chemicals. These patients report little diminution in their cognitive abilities as a result. It seems that a substantial core of cognitive abilities remain despite the patients being all but severed from meaningful interaction with the wider world. As a result, Aizawa questions whether distal or historical factors could actively contribute to these elements of ongoing cognition---in the way scratch paper contributes to division or rotating Tetris blocks on screen contribute to mental rotation. Of course, locked-in patients would not be able to engage in many of the  perceptual/cognitive processes (like Tetris-playing) that involve active interaction with the environment. However, here the relevant external factors are nearby in time and space, so this lends little support to the view that distal or historical factors play a causal role in ongoing cognition. 

More plausible roles for such factors are constitutive and explanatory. For example, consider the frog thought experiment discussed in the previous section. Here we appealed to the present and historical ecological niche of the frog species. Facts about these niches go beyond the immediate environment of any particular frog, but plausibly, they contribute to fixing the meaning of symbols in individual frog's brains. Further, they allow us to understand the biological significance of cognitive and perceptual processes within frogs. If we gave a frog to alien scientists without giving any information about its environment or evolutionary history, we would significantly inhibit their understanding of frog cognition and behavior. In other words, it seems quite plausible that distal or historical factors are relevant to a semantic interpretation of mental symbols, and these interpretations are of vital importance for understanding and explaining ongoing cognitive processes.

\subsection{Embodiment, Learning, and Artificial Intelligence}

Besides these broader themes, there were several more specific themes of note. For example, the importance of learning and optimization over multiple timescales arose in both talks and discussions. In human development, we have already seen that infants' brains and bodies develop over significant periods of time and within these developmental periods, shorter episodes of learning take distinctive forms and feed back into intermediate-term patterns of exploration that facilitate further learning. In the context of AI systems, Nick Cheney and Tyler Marghetis discussed the importance of optimizing brains and bodies on different time scales, and Yoshua Bengio argued that neural networks should mimic both fast ``System 1'' style thinking and slower ``System 2'' style thinking, with the latter enabling the rapid redeployment of specialized modules for novel tasks. 

Of course, learning over multiple timescales is but one of the elements of human learning that participants found lacking in today's large neural network models. While there was some optimism about the future prospect for existing approaches to designing and training neural networks, even the optimists admitted that progress might well require data from multiple modalities (even if exploration and embodiment, per se, are not required). That said, getting this kind of rich data is much more natural for embodied agents and there was little resistance to making learning for AI systems more like learning in embodied agents or to the idea that doing so might help AI systems to overcome some of their characteristic failure modes. Hence, while participants differed on what what necessary \textit{in principle}, they more often agreed about what might useful for AI in practice.

There are many well-documented failure modes of today's neural networks, but the propensity for large language models to value ``truthiness'' was especially emphasized. The idea here is that large language models seem to prioritize generating human-like text based on their sophisticated ability to learn distributional information about natural language from large corpora of text. This aim is not orthogonal to truth, but producing true statements seems at best a means to the end of producing human-like text. The truth generally matters to people, honest or otherwise, and it is a very different kind of discourse that proceeds without regard for it. The willingness of large language models to confidently confabulate, contradict themselves, or generate an arbitrary admixture of truth and falsehood suggest that they ``communicate'' in a very different way than their human interlocutors. There was some interesting speculation at the workshop about why this might be the case. One plausible idea is that AI systems don't have a stake in a world in which they are embedded. As Chris Potts suggested, there may be much that an AI system can learn from observing a huge number of examples of chess, but it is much less clear that such a system would be good at chess whatever else it might know about it. Producing likely chess moves and producing good chess moves are different things. Similarly, agents with practical or communicative goals will be concerned with what is and is not true (even if they sometimes have reason to lie). This comes quite naturally to evolved agents, such as humans, who have been selected for having fitness-enhancing goals and desires, but it is something notably lacking in many contemporary AI systems. 

\subsection{Future Directions}

A striking commonality among the diverse talks at the workshop was the lack of any commitment to a specific approach to developing machine intelligence. For example, even those most optimistic about the prospects of existing systems envisioned substantial changes to their architecture, optimization, evaluation, or training data. Understanding intelligence and building intelligent systems are among the hardest problems that humans face. Nevertheless, we are surrounded by evolved systems that excel at learning, problem solving, and communication. All of these systems are embodied and situated, entering into complex interactions between mind, body, and environment. There is little doubt that better understanding these interactions will further our understanding of intelligence in both natural and artificial systems.

\newpage
\section*{Acknowledgments}
The workshop was funded by a grant from the National Science Foundation (\#2020103) as part of the Foundations of Intelligence in Natural and Artificial Systems project at the Santa Fe Institute.

\section{References}

\begin{flushleft}

\begin{list}{}
{\leftmargin=1em \itemindent=-1em \itemsep=-.4em}

\item Burge, T. (1979). \textit{Individualism and the mental}. \textit{Midwest studies in philosophy}, \textit{4}(1), 73-122.

\item Clark, A., \& Chalmers, D. (1998). The extended mind. \textit{Analysis}, \textit{58}(1), 7-19.

\item Dennett (1990). Evolution, error, and intentionality. In D. Partridge \& Y. Wilks (Eds.), \textit{The foundations of artificial intelligence: A sourcebook}. Cambridge, UK: Cambridge University Press. 190-212.

\item Evans, J. S. B., \& Stanovich, K. E. (2013). Dual-process theories of higher cognition: Advancing the debate. \textit{Perspectives on psychological science}, \textit{8}(3), 223-241.

\item Frankfurt (1986). On bullshit. \textit{Raritan quarterly review}. \textit{6}(2). 81–100. 

\item Harnad, S. (1990). The symbol grounding problem. \textit{Physica D: Nonlinear phenomena}, \textit{42}(1-3), 335-346.

\item Kahneman, D. (2011). \textit{Thinking fast and slow}. New York, NY: Farrar, Straus, and Giroux.

\item Lettvin, J. Y., Maturana, H. R., McCulloch, W. S., \& Pitts, W. H. (1959). What the frog's eye tells the frog's brain. \textit{Proceedings of the IRE}, \textit{47}(11), 1940-1951.

\item Millikan, R. (1989) Biosemantics. \textit{The journal of philosophy}. \textit{86}(6). 281-297.

\item Pearl, J. (2000). \textit{Causality}. Cambridge, UK: Cambridge University Press.

\item Putnam, H. (1974). Meaning and reference. \textit{The journal of philosophy}, \textit{70}(19), 699-711.

\item Woodward, J. (2003). \textit{Making things happen: A theory of causal explanation}. Oxford, UK: Oxford University Press.

\end{list}
\end{flushleft}

\end{document}